%% file: main_arxiv.tex
\documentclass{article}
\usepackage{iclr2027_conference,times}
\input{math_commands.tex}

\usepackage{amsmath}
\usepackage{booktabs}
\usepackage{array}
\usepackage{float}
\usepackage{graphicx}
\usepackage{microtype}
\usepackage{xcolor}

\usepackage{hyperref}
\usepackage{url}
\newcommand{\bench}{\textbf{VLA-HWBench}}
\usepackage{pifont}
\newcommand{\cmark}{\ding{51}}
\newcommand{\xmark}{\ding{55}}

\graphicspath{{figs/}}
\usepackage{tcolorbox}
\usepackage{enumitem}
\definecolor{DraftAmber}{RGB}{145,77,0}
\definecolor{SlotBackground}{RGB}{245,247,249}
\definecolor{SlotBorder}{RGB}{158,172,185}
\definecolor{FindingBlue}{RGB}{27,58,91}
\definecolor{FindingBackground}{RGB}{242,247,251}
\definecolor{MethodRowGray}{RGB}{238,238,238}
\newtcolorbox{finding}[2]{colback=FindingBackground,colframe=FindingBlue,
  colbacktitle=FindingBlue,coltitle=white,fonttitle=\bfseries,
  title={Finding #1: #2},arc=0pt,boxrule=0.5pt,boxsep=0pt,
  left=6pt,right=6pt,top=5pt,bottom=5pt,toptitle=4pt,bottomtitle=4pt,
  before skip=7pt,after skip=7pt}
\newtcolorbox{insight}[2]{colback=FindingBackground,colframe=FindingBlue,
  colbacktitle=FindingBlue,coltitle=white,fonttitle=\bfseries,
  title={Insight #1: #2},arc=0pt,boxrule=0.5pt,boxsep=0pt,
  left=6pt,right=6pt,top=5pt,bottom=5pt,toptitle=4pt,bottomtitle=4pt,
  before skip=7pt,after skip=7pt}

\newtcolorbox{insightbox}{colback=FindingBackground,colframe=FindingBlue,
  arc=0pt,boxrule=0.5pt,boxsep=0pt,
  left=6pt,right=6pt,top=5pt,bottom=5pt,
  before skip=7pt,after skip=7pt}

\title{VLA-HWBench: Where the Time Goes When\\VLA Models Run on Edge Hardware}

\author{
\textbf{Borui Xu\textsuperscript{1},
Hongshi Tan\textsuperscript{1},
Wenqi Jiang\textsuperscript{1},
Bingsheng He\textsuperscript{1}} \\
\textbf{{} Ming Wu\textsuperscript{2},
Jason Zeng\textsuperscript{2},
Michael Heinrich\textsuperscript{2}} \\
\textsuperscript{1}National University of Singapore \\
\textsuperscript{2}0G Labs
}

\iclrfinalcopy

\begin{document}
\maketitle
\suppressfloats[t]  

\begin{abstract}

{Vision--language--action (VLA) models increasingly run on edge devices aboard robots, where inference latency constrains closed-loop robot control. However, it remains unclear how to design VLA models, their inference software and edge hardware so that robots can run VLA inference in real time.} Most VLA models report only task success, and {prior studies cover few of the ways recent VLAs generate actions, split a call into at most three phases and do not compare inference optimizations on the same models.} Therefore, we introduce \bench, a benchmark that measures 18 open VLA models on their released software and on alternative backends, on an NVIDIA Jetson AGX Thor and an AMD Ryzen AI Max+ PRO 395, with a seven-stage breakdown of each call. {Our study reveals three insights. First, {where the time goes depends on how a model generates actions, not on its size. Action generation is the largest stage in 14 of the 19 configurations on the Jetson,} latency spans $82\times$ on the Jetson and $157\times$ on the AMD platform, {and parameter count explains at most 15\% of its variance. Models} that generate long action chunks in a few flow-matching steps keep up with control{,} whereas token-by-token decoding does not. Second, kernel fusion with 4-bit weights helps most{, running} $\pi_{0.5}$ and GR00T N1.7 2.6--2.9$\times$ faster than PyTorch with \texttt{torch.compile}, but it exists for only these two models and only on the Jetson. {Optimization also moves the time to the stages it shortens least, as CPU preprocessing grows from 12\% to 39\% of the GR00T N1.7 call.} Third, after optimization, memory bandwidth limits only the stages that process few tokens on the Jetson, such as the denoising steps, whereas software limits all other stages, from unfused kernels to {serial preprocessing on the host CPU.} Overall, low-latency edge VLAs should use few action-generation iterations, inference software should extend kernel fusion and low-bit weights to more models and platforms, and edge hardware needs high memory bandwidth and a fast host CPU.}

\end{abstract}

\input{sections/01_introduction}

\input{sections/02_related}
\input{sections/03_study}
\input{sections/04_design}

\input{sections/05_results}
\input{sections/06_conclusion}

\clearpage


\bibliography{references}
\bibliographystyle{iclr2027_conference}

\clearpage
\input{sections/appendix}

\end{document}

%% file: math_commands.tex
\usepackage{amsmath,amsfonts,bm}

\def\eqref#1{equation~\ref{#1}}

\def\1{\bm{1}}

\DeclareMathAlphabet{\mathsfit}{\encodingdefault}{\sfdefault}{m}{sl}
\SetMathAlphabet{\mathsfit}{bold}{\encodingdefault}{\sfdefault}{bx}{n}



%% file: sections/01_introduction.tex
\section{Introduction}
\label{sec:intro}

Vision--language--action (VLA) models have emerged as a promising approach to general-purpose robotic control, mapping visual observations and language instructions directly to robot actions~\citep{brohan2023rt2,kim2024openvla,octo2024,black2024pi0,groot2025}. Unlike text or image generation, VLA inference operates within a continuously evolving physical system: the robot may continue executing previous commands while objects or obstacles change state or position~\citep{falanga2019howfast,morrison2018ggcnn}, causing observations to become stale before the corresponding actions are executed. Practical deployment therefore increasingly calls for low-latency inference close to the robot. This is particularly important when wireless connectivity is variable or intermittent, or when privacy-sensitive camera observations cannot be transmitted to remote servers. Meanwhile, advances in edge compute hardware~\citep{jetsonthor,strixhalo} are making on-robot VLA inference increasingly feasible.


However, the observation-to-action characteristics of state-of-the-art VLA models on edge devices have not been systematically characterized. {Recent studies have begun to characterize VLA inference through measurements on edge and datacenter GPUs \citep{crossplatform2025,vlaxpu2026}, profiling of a single architecture \citep{vishwanathan2026characterizing} and analytical performance models \citep{jiang2026vlaperf}, but three gaps remain.}
\textbf{First}, {they cover few of the ways in which recent VLAs generate actions. Recent models differ in architecture and training and denoise action chunks by flow matching, decode them in parallel, generate them token by token or reason before acting. These choices change how much sequential work a call performs, but the measured studies cover at most five models and two of these four forms.}
\textbf{Second}, {they divide a call into at most three phases, which cannot show where the time goes. CPU preprocessing, data transfer, vision encoding, prefill and action generation stress different resources, and none of these studies times CPU preprocessing.}
\textbf{Third}, {they do not compare inference optimizations on the same models. None of the four studies evaluates a specialized backend and only one tests low-bit weights, while each backend is evaluated only on the models it targets \citep{tensorrt,flashrt,embodiedcpp2026,vlacpp2026}. It thus remains unclear which optimization helps most, which stages it shortens and whether its gain carries over to other models and platforms.}

To bridge these gaps and provide a systematic basis for practical on-robot deployment, we present {\bench}, a benchmark for characterizing the complete observation-to-action path of state-of-the-art VLA models on edge devices. \bench{} is designed to incorporate additional VLA models, inference backends, and edge hardware as the ecosystem evolves.

{First, to cover the forms of action generation, \bench{} evaluates \textbf{18} widely used open VLA models spanning all four forms, selected based on Hugging Face downloads and community stars, on two architecturally distinct edge platforms, the NVIDIA Jetson AGX Thor and AMD Ryzen AI Max+ PRO 395. Each model runs in a common reference configuration using eager PyTorch, with optional optimizations disabled, on the same LIBERO-10 observation~\citep{liu2023libero} and timing boundary. Second, to show where the time goes, \bench{} divides each call into the seven stages of Figure~\ref{fig:stages}, timed at the stage boundaries and split by kernels inside CUDA graphs. Third, to study optimizations systematically, \bench{} enables compilation, specialized backends (TensorRT and FlashRT) and lower precision one at a time on two representative models and attributes each gain to stages.} With \bench{}, we answer three questions.

\begin{figure}[t]
  \centering
  \includegraphics[width=\linewidth]{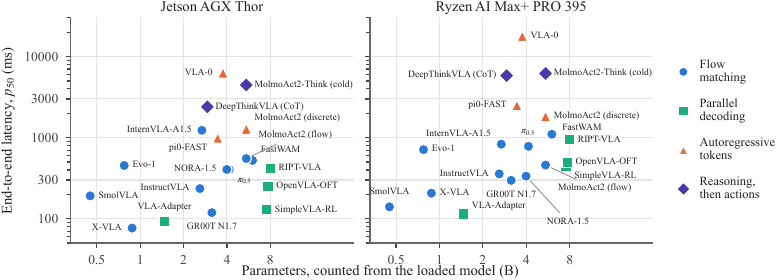}
  \caption{\textbf{End-to-end latency versus loaded parameter count} on the Jetson AGX Thor (left) and the Ryzen AI Max+ PRO 395 (right), with one point per reference configuration.}
  \label{fig:params}
  \vspace{-2mm}
\end{figure}

\textbf{(Q1) {Where does the time go in VLA models, and what sets their latency?}} Without optional optimizations, {action generation is the largest stage in 14 of the 19 configurations on the Jetson, and} latency ranges from 77\,ms to 6.3\,s on the Jetson ($82\times$) and from 114\,ms to 17.9\,s on the AMD platform ($157\times$), {with} only 10 and 9 of the 19 configurations generating actions as fast as LIBERO executes them. As Figure~\ref{fig:params} shows, {a least-squares fit of log latency on log parameter count explains at most 15\% of the variance, and latency instead follows the number of iterations of action generation.}

\textbf{(Q2) {How much do optimizations help, and where does the time move?}} {On the Jetson, compilation, a specialized backend and lower precision make $\pi_{0.5}$ $2.5\times$, $1.8\times$ and $1.6\times$ faster in turn, $7.2\times$ in total, and GR00T N1.7 $1.02\times$, $1.4\times$ and $1.8\times$, $2.6\times$ in total. A backend with kernel fusion and 4-bit weights gives the largest gain, 2.6--2.9$\times$ over compilation, but exists only for these two models and only on the Jetson. {Optimization also moves the time to the stages it shortens least, as CPU preprocessing grows from 12\% to 39\% of the GR00T N1.7 call.}}


\textbf{(Q3) {What limits the remaining time after optimization?}} {{No single resource limits the remaining time.} Action generation and other stages with few tokens are memory-bound and on the Jetson already run at more than half of the peak memory bandwidth{,} whereas vision encoding and long prefills reach only 6--42\% of what the hardware allows because of unfused kernels and missing low-precision support, and CPU preprocessing is limited by serial Python and idle-thread wake-ups{.}}


Overall, the latency of a VLA on edge hardware is determined jointly by its action-generation method, its runtime backend and the platform rather than by its model size. For future designs, edge VLAs should therefore reduce action-generation iterations, for example through fewer denoising steps or parallel decoding, rather than simply shrink the model; inference software should extend kernel fusion and low-bit execution, which provide the largest gains, to more models and platforms while optimizing host preprocessing as carefully as GPU execution; and edge hardware should prioritize memory bandwidth and a fast host CPU over peak GPU compute.

%% file: sections/02_related.tex
\begin{figure}[t]
  \centering
  \includegraphics[width=\linewidth]{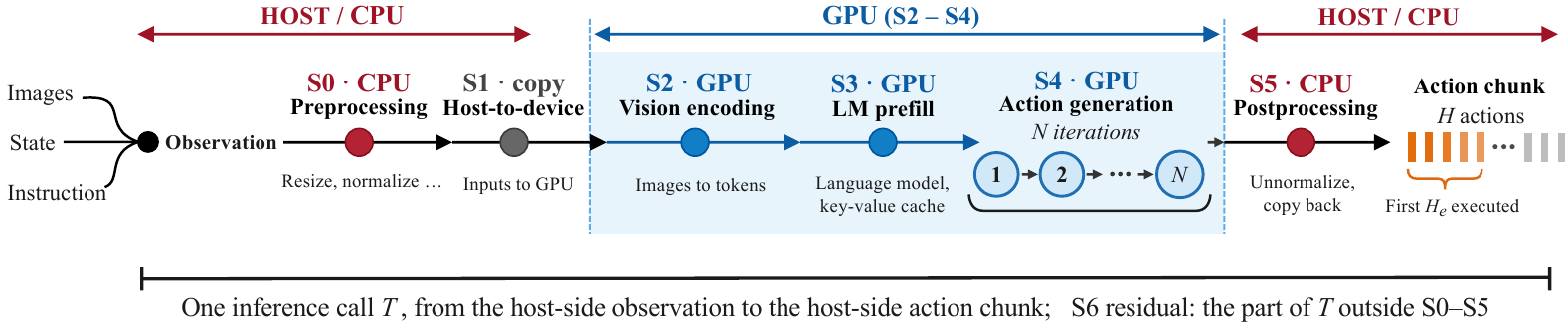}
  \caption{\textbf{The seven stages of one inference call.} S0 and S5 run on the CPU and S2--S4 on the GPU.}
  \label{fig:stages}
  \vspace{-2mm}
\end{figure}

\section{Related Work}
\label{sec:related}

\textbf{VLA models and action generation.} Released VLA models \citep{brohan2023rt2,kim2024openvla,octo2024,black2024pi0,groot2025} generate an action chunk with a flow-matching head that integrates an action expert over a fixed number of steps \citep{lipman2022flow,black2024pi0}, by parallel decoding in one pass \citep{kim2025oft}, token by token \citep{brohan2023rt2,kim2024openvla,pertsch2025fast}, or after a reasoning trace \citep{deepthinkvla,molmoact2}. They are compared by task success; \bench{} measures what each form of action generation costs at inference time on edge hardware.


\textbf{VLA inference on edge devices.} \citet{crossplatform2025} time five models on a Jetson AGX Orin and four datacentre GPUs, \citet{vlaxpu2026} profile {seven policies, five of them VLAs,} on accelerators including the Jetson Thor and find the action expert memory-bound, \citet{vishwanathan2026characterizing} attribute up to 75\% of the latency on edge architectures to action generation, and VLA-Perf \citep{jiang2026vlaperf} models $\pi_0$-like architectures analytically. Table~\ref{tab:related} compares \bench{} with prior studies in terms of model coverage, action-generation stages, hardware platforms, inference backends, and weight precision.

\begin{table}[t]
  \centering
  \caption{{\textbf{Comparison with prior VLA edge-inference studies.}}}
  \label{tab:related}
  \footnotesize
  \setlength{\tabcolsep}{3pt}
  \begin{tabular}{@{}lccccccccc@{}}
    \toprule
    Study & Models & Forms & Measured & Non-NVIDIA & Stages & CPU & Backend & Low-bit & Real \\
     & & & & edge & & prep. & & & input \\
    \midrule
    \citeauthor{crossplatform2025} & 5 & 2 & \cmark & \xmark & -- & \xmark & \xmark & \xmark & \xmark \\
    \citeauthor{vlaxpu2026} & 5 & 2 & \cmark & \cmark & 3 & \xmark & \xmark & \cmark & \xmark \\
    \citeauthor{vishwanathan2026characterizing} & 1 & 1 & \cmark & \xmark & 3 & \xmark & \xmark & \xmark & \xmark \\
    VLA-Perf & -- & 3 & \xmark & \xmark & 3 & \xmark & \xmark & \xmark & \xmark \\
    \midrule
    \bench{} & 18 & 4 & \cmark & \cmark & 7 & \cmark & \cmark & \cmark & \cmark \\
    \bottomrule
  \end{tabular}
  \vspace{-2mm}
\end{table}

\textbf{Inference runtimes and optimizations.} TensorRT \citep{tensorrt} and FlashRT \citep{flashrt} use fused kernels and CUDA graphs, Embodied.cpp \citep{embodiedcpp2026} and vla.cpp \citep{vlacpp2026} are C++ runtimes for selected models, and LeRobot exposes \texttt{torch.compile} \citep{ansel2024pytorch2}. Lower precision cuts memory traffic and uses faster arithmetic \citep{dettmers2022llmint8,frantar2023gptq,lin2024awq}, and compressed action tokens \citep{pertsch2025fast}, fewer denoising steps \citep{salimans2022progressive,song2023consistency}, real-time chunking \citep{black2025rtc} and asynchronous inference \citep{shukor2025smolvla,yang2026jetsonpi} reduce or hide the work behind each chunk. Each is evaluated on the models it targets; \bench{} applies the options of released models stage by stage on fixed weights and compares their gains across the models and the two platforms.

%% file: sections/03_study.tex
\section{Studying VLA Inference Latency on Edge Devices}
\label{sec:study}

\subsection{Research Problem}
\label{sec:problem}

A VLA model maps an observation of camera images, the robot's proprioceptive state and a language instruction to a chunk of $H$ actions that the controller executes before it calls the model again. We denote by $T$ the latency of one inference call, measured from the host-side observation to the host-side action chunk, and divide it into the seven stages S0--S6 of Figure~\ref{fig:stages},
\begin{equation}
  T = \sum_{k=0}^{6} T_k,
  \label{eq:stages}
\end{equation}
where $T_k$ is the time of stage S$k$, namely CPU preprocessing (S0), the host-to-device copy (S1), vision encoding (S2), the prefill pass of the language model (S3), action generation (S4), postprocessing (S5), and the residual that the other stages do not cover, such as Python dispatch between them (S6). Action generation runs $N$ iterations one after another, so that
\begin{equation}
  T_4 \approx N\, t,
  \label{eq:passes}
\end{equation}
where $t$ is the cost of one iteration. $N$ is the number of denoising steps for a flow-matching head, one for parallel decoding, the number of generated tokens for an autoregressive head, and the number of reasoning tokens plus action iterations for a reasoning model.

\subsection{Metrics}
\label{sec:metrics}

To systematically answer the three research questions, we define a corresponding set of metrics, each targeting the key property examined by one question.

\textbf{Q1, {where the time goes}.} The first question compares {the stage shares $T_k/T$,} the latency $T$ and the inference time per action $T/H$ across models in their reference configurations, which run each model without optional optimizations. LIBERO's simulator runs its controller at 20\,Hz \citep{liu2023libero,robosuite}, so it executes one action every 50\,ms, and a model {keeps up} only if $T/H < 50$\,ms.


\textbf{Q2, optimizations.} The second question measures the speed-up $T_0/T$ of an optimization, where $T_0$ is the latency of the reference configuration and $T$ the latency with the optimization. It also compares the actions before and after each compilation and precision step.


\textbf{Q3, bottlenecks.} {The third question asks what limits the fastest configurations, using the roofline model \citep{williams2009roofline}. A GPU stage with $F$ floating-point operations and $W$ bytes of weights takes at least its compute bound $F/P$ or its memory bound $W/B$, whichever is larger, where $P$ is the peak matrix-unit throughput in its number format and $B$ the peak memory bandwidth. It is memory-bound if $F/W$ lies below the ridge point $P/B$. A stage close to its bound is limited by the hardware, and one far from it by its software.}

\subsection{Design Principles}
\label{sec:principles}

\textbf{Consistent runtime backends.} We run each model through the inference entry point of its own LIBERO evaluation with the library versions its authors pin, compare models only in their reference configurations, and add optimizations one at a time on fixed weights and input.

\textbf{Stage-level attribution.} We synchronize the device at every stage boundary of each release's own code, check that the stages add up to the end-to-end latency, and separate the stages that a runtime captures in one CUDA graph by the graph's kernels.

\textbf{Two edge platforms.} Two platforms with Arm and x86 hosts, which differ by $4.4\times$ in peak 16-bit throughput but only $1.1\times$ in memory bandwidth, separate the effects of host, GPU and memory.

%% file: sections/04_design.tex
\section{\bench{} Benchmark Design}
\label{sec:design}

\subsection{Model Coverage}
\label{sec:models}

{We select 18 open VLA models that were released in 2025 or 2026, report strong LIBERO success, publish a LIBERO checkpoint with its evaluation code and are widely used, and that together cover the four main forms of action generation. Table~\ref{tab:models} lists them, and Appendix Table~\ref{tab:models-pop} gives the GitHub stars of their code and the downloads of their checkpoints.} Each model runs its LIBERO-10 checkpoint with the settings of its own LIBERO evaluation, which Appendix Table~\ref{tab:models-full} details, and MolmoAct2's two action heads give 19 reference configurations.

\begin{table}[t]
  \centering
  \caption{The 18 models, with the parameters and weight types of the loaded model and the chunk size $H$ of each release's LIBERO evaluation. $^\dagger$Runs on LeRobot, the others on the authors' code.}
  \label{tab:models}
  \small
  \setlength{\tabcolsep}{3.6pt}
  \renewcommand{\arraystretch}{0.95}
  \input{tables/models.tex}
  \vspace{-3mm}
\end{table}

\subsection{Runtime Backends}
\label{sec:paths}

The models run on four runtime backends.
\begin{itemize}[leftmargin=*,itemsep=0pt,topsep=2pt]
  \item \textbf{LeRobot 0.6.1} \citep{lerobot,lerobot2026} is a library for robot learning that runs each policy as a PyTorch program and offers \texttt{torch.compile} as an option.
  \item \textbf{The authors' code} is the inference entry point of a release's own LIBERO evaluation, which runs the model as a PyTorch program with the library versions the authors pin.
  \item \textbf{FlashRT} \citep{flashrt} is an inference runtime for VLA models that replaces the PyTorch program with fused kernels in captured CUDA graphs and offers FP8 and NVFP4 weights.
  \item \textbf{TensorRT} \citep{tensorrt} compiles each part of a model into an optimized inference engine.
\end{itemize}
Q1 runs every model in its reference configuration, on LeRobot or the authors' code with every optional optimization off. Q2 adds the optimizations of all four backends one at a time to the reference configurations of $\pi_{0.5}$ and GR00T N1.7, and Q3 analyzes the fastest configurations of Q2. FlashRT and TensorRT run only on the Jetson, and Appendix~\ref{app:q2} reports the options of the other models.

\subsection{Platforms}
\label{sec:platforms}

\begin{table}[t]
  \centering
  \caption{Specifications of the two platforms. $^{a}$Derived from the CU count and clock.}
  \label{tab:platforms}
  \small
  \input{tables/platforms.tex}
  \vspace{-2mm}
\end{table}

We run the benchmark on two edge platforms, an NVIDIA Jetson AGX Thor and an AMD Ryzen AI Max+ PRO 395, each with a GPU that shares memory with the CPU. Table~\ref{tab:platforms} lists their specifications.

\subsection{Measurement Protocol}
\label{sec:protocol}

\textbf{Input and timing.} Every configuration processes the same real LIBERO-10 observation \citep{liu2023libero}, two $256\times256$ camera images, the robot state and the task instruction, adapted only where a release's input contract requires it. Each configuration runs in its own process with 20 warm-up calls and up to 100 timed calls, trimmed to fit 90\,s but never below 20, each bracketed by device synchronization, and we report the median latency. To check how the content of the input affects latency, we also run every configuration on 24 different frames of one episode.

\textbf{Stage attribution.} Synchronized timers at the stage boundaries of each model's own software measure the stages of Equation~\ref{eq:stages}, which Appendix Table~\ref{tab:stagedefs} details. The reference configurations run without CUDA graphs, so each boundary that a release's code exposes is timed directly, including the vision-encoder and prefill calls inside LeRobot's policies, and a stage that the code runs inside another one is counted in that one. We get the cost $b$ of one denoising step or camera view by fitting latency as $a + bN$ over their number $N$, and we split the CUDA graphs of the optimized configurations into stages by the order of their kernels and how often each repeats, once per step or layer. 


\textbf{Profiling.} {The profiler in Pytorch traces the GPU kernels of the optimized configurations, which we sort into GEMM or attention kernels on the matrix units, GEMMs without them and other kernels, and records the shapes and number formats of every operator in an eager run of the same workload, which give $F$ and $W$ of each stage. For CPU preprocessing, we time its sub-steps, vary the number of CPU threads and read the cores' idle-state counters.}

%% file: tables/models.tex
\begin{tabular}{@{}llrlr@{}}
\toprule
Action generation & Model & Params (B) & Weights & Action chunk $H$ \\
\midrule
Flow matching & X-VLA~\citep{xvla} & 0.88 & BF16 & 30 \\
 & GR00T N1.7~\citep{groot2025} & 3.14 & BF16 & 16 \\
 & SmolVLA$^\dagger$~\citep{shukor2025smolvla} & 0.45 & BF16 + FP32 & 50 \\
 & InstructVLA~\citep{instructvla} & 2.60 & BF16 + FP32 & 8 \\
 & $\pi_{0.5}$$^\dagger$~\citep{pi05} & 4.14 & BF16 + FP32 & 50 \\
 & NORA-1.5~\citep{nora15} & 3.99 & BF16 & 5 \\
 & Evo-1~\citep{evo1} & 0.78 & BF16 + FP32 & 50 \\
 & FastWAM$^\dagger$~\citep{fastwam} & 6.02 & BF16 & 32 \\
 & MolmoAct2 (flow)$^\dagger$~\citep{molmoact2} & 5.44 & BF16 & 10 \\
 & InternVLA-A1.5~\citep{internvla} & 2.68 & BF16 + FP32 & 8 \\
\midrule
Parallel & VLA-Adapter~\citep{vlaadapter} & 1.47 & BF16 & 8 \\
decoding & SimpleVLA-RL~\citep{simplevlarl} & 7.54 & BF16 & 8 \\
 & OpenVLA-OFT~\citep{kim2025oft} & 7.71 & BF16 & 8 \\
 & RIPT-VLA~\citep{ript} & 7.99 & BF16 + FP32 & 8 \\
\midrule
Autoregressive & pi0-FAST$^\dagger$~\citep{pertsch2025fast} & 3.45 & BF16 + FP32 & 10 \\
tokens & MolmoAct2 (discrete)$^\dagger$~\citep{molmoact2} & 5.44 & BF16 & 10 \\
 & VLA-0~\citep{vla0} & 3.75 & BF16 & 8 \\
\midrule
Reasoning, & DeepThinkVLA (CoT)~\citep{deepthinkvla} & 2.92 & BF16 & 10 \\
then actions & MolmoAct2-Think (cold)~\citep{molmoact2} & 5.45 & BF16 & 10 \\
\bottomrule
\end{tabular}

%% file: tables/platforms.tex
\begin{tabular}{@{}lll@{}}
\toprule
& Jetson AGX Thor (T5000) & Ryzen AI Max+ PRO 395 \\
\midrule
GPU & Blackwell, 2560 CUDA cores, 1.57\,GHz & RDNA 3.5, 40 CUs, 2.9\,GHz \\
CPU & 14 Neoverse-V3AE cores, 2.6\,GHz & 16 Zen 5 cores, 5.1\,GHz \\
Memory & 128\,GB LPDDR5X, 273\,GB/s & 128\,GB LPDDR5X-8000, 256\,GB/s \\
FP16 / FP32 (TFLOP/s) & 258 / 8.1 & 59.4$^{a}$ / 29.7$^{a}$ \\
FP8 / FP4 (TFLOP/s) & 517 / 1035 & -- \\
Power & 40--130\,W & 45--120\,W \\
Software & CUDA 13.2, PyTorch 2.11 & ROCm 7.2, PyTorch 2.14 \\
\bottomrule
\end{tabular}

%% file: sections/05_results.tex

\section{Benchmark Results}
\label{sec:results}

Sections~\ref{sec:q1}--\ref{sec:q3} trace where the time goes before, during and after optimization.

\begin{figure}[t]
  \centering
  \includegraphics[width=\linewidth]{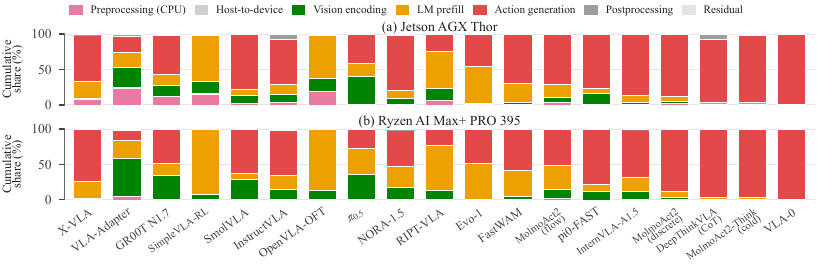}
  \caption{\textbf{Stage breakdown of the 19 reference configurations} on the Jetson (a) and the AMD platform (b), with a fused stage counted in the stage that contains it.}
  \label{fig:q1stages}
  \vspace{-2mm}
\end{figure}

\begin{figure}[]
  \centering
  \includegraphics[width=\linewidth]{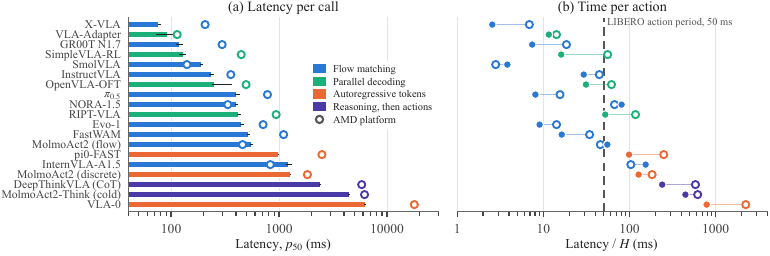}
  \caption{\textbf{Latency of the 19 reference configurations.} (a) Median latency on the Jetson (bars, 5th--95th percentile) and the AMD platform (circles). (b) Latency over the chunk size $H$.}
  \label{fig:q1}
  \vspace{-2mm}
\end{figure}

\subsection{Where Does the Time Go in VLA Models, and What Sets Their Latency?}
\label{sec:q1}

In most reference configurations, action generation takes the largest share of the call, and latency, which spans up to two orders of magnitude, follows the iterations of action generation rather than model size, and only about half of the configurations keep up with LIBERO's control rate.

\textbf{Where the time goes.} Figure~\ref{fig:q1stages} shows that action generation is the largest stage in 14 of the 19 configurations on the Jetson, with 42--99\% of the call, as in prior work (up to 75\%, \citealp{vishwanathan2026characterizing}). The five exceptions, the four parallel decoders and Evo-1, spend more of the call in prefill or vision encoding than in action generation, as does $\pi_{0.5}$ on the AMD platform, whose FP32 vision tower encodes three image slots, one empty.

\textbf{Latency range.} Figure~\ref{fig:q1}a shows that latency ranges from 77\,ms for X-VLA to 6.3\,s for VLA-0 on the Jetson ($82\times$) and from 114\,ms to 17.9\,s on the AMD platform ($157\times$), with a similar ranking (Spearman's $\rho = 0.91$). Releases generate 5 to 50 actions per call, and Figure~\ref{fig:q1}b shows that long chunks hide slow calls, as $\pi_{0.5}$ spends less time per action than VLA-Adapter though its call takes $4.4\times$ longer. Only 10 of the 19 configurations on the Jetson and 9 on the AMD platform keep up with LIBERO's control rate.

\textbf{Model size.} In Figure~\ref{fig:params}, a least-squares fit of log latency on log parameter count, one point per checkpoint, explains only 9\% of the variance on the Jetson and 15\% on the AMD platform. GR00T N1.7 and VLA-0 hold 3.1 and 3.8\,B parameters, yet on the Jetson VLA-0 is $53\times$ slower, and OpenVLA-OFT holds $5.2\times$ the parameters of VLA-Adapter, also a parallel decoder, but is only $2.7\times$ slower per call.

\textbf{{Iterations.}} Equation~\ref{eq:passes} accounts for the gap. On the Jetson, one flow-matching step costs 4--107\,ms and parallel decoding needs a single {iteration}, whereas autoregressive heads pay one language-model pass of 20--52\,ms per token, so the 208 decimal tokens of VLA-0 take 99\% of its call and the reasoning tokens of DeepThinkVLA and MolmoAct2-Think about 91\%. On identical weights, MolmoAct2's discrete head is $2.3\times$ slower than its flow head, and the four configurations with scene-dependent token counts vary by 12--122\% across 24 frames.

\begin{insightbox}
\textbf{Insight 1.} In most of the 19 reference configurations, \textbf{the time goes mainly to action generation, and latency follows the number of its iterations rather than the parameter count}, so token-by-token decoding and reasoning cannot keep up with LIBERO's 20\,Hz control whereas flow matching over long action chunks can. \textbf{Future edge VLAs should cut these iterations}, with fewer denoising steps, parallel decoding or longer action chunks, rather than shrink the model.
\end{insightbox}

\begin{figure}[t]
  \centering
  \includegraphics[width=\linewidth]{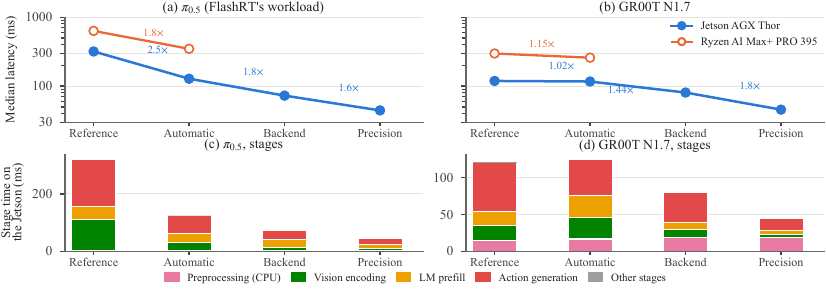}
  \caption{\textbf{Optimizations applied in stages to $\pi_{0.5}$ and GR00T N1.7.} (a, b) Latency of the fastest configuration with the optimizations up to each stage, labeled with the speed-up of the stage. (c, d) Their stage times on the Jetson.}
  \label{fig:q2}
  \vspace{-2mm}
\end{figure}

\subsection{How Much Do Optimizations Help, and Where Does the Time Move?}
\label{sec:q2}

We optimize $\pi_{0.5}$ and GR00T N1.7, the two models for which all three levels exist, and Figure~\ref{fig:q2} shows the levels from the least to the most invasive. Automatic optimization compiles the unchanged model, with LeRobot's compile option for $\pi_{0.5}$ and Isaac-GR00T's compile recipe for GR00T N1.7. A specialized backend replaces the runtime, with FlashRT for both models and TensorRT for GR00T N1.7, and lower precision stores the weights in FP8 or NVFP4 inside FlashRT. So that every step changes one thing, all $\pi_{0.5}$ configurations run FlashRT's workload, 10 actions and two image slots.

\textbf{Automatic optimizations.} Compilation helps only when it captures the whole model. LeRobot's compile option captures the whole $\pi_{0.5}$ call in one CUDA graph, cutting its kernel launches from 13k to 15, and also switches {FP32} matrix multiplications to TF32, making the call $2.5\times$ faster on the Jetson and $1.8\times$ on the AMD platform, which has no TF32. Isaac-GR00T's recipe compiles only the action head, which speeds that stage up by $1.4\times$ but slows the uncompiled vision encoding and prefill, so the call gains only $1.02\times$ on the Jetson. Compilation also changes $\pi_{0.5}$'s actions and can flip a gripper command, as Appendix~\ref{app:q2} details.

\textbf{Backend replacement.} FlashRT makes $\pi_{0.5}$ $1.8\times$ and GR00T N1.7 $1.4\times$ faster than the level before, whereas TensorRT gains less for GR00T N1.7, as it keeps the vision encoder in {FP32}. A backend also changes the number format, since FlashRT runs $\pi_{0.5}$'s FP32 layers in {FP16}.

\textbf{Lower precision.} FP8 and NVFP4 make both models another 1.3--1.8$\times$ faster, mostly in action generation. FP8 keeps the cosine similarity to the FP16 actions at 0.998 or higher and NVFP4 at 0.93 or higher. In total, $\pi_{0.5}$ falls from 318 to 44\,ms on the Jetson and GR00T N1.7 from 119 to 46\,ms.

\textbf{Where the time moves.} Figure~\ref{fig:q2}c,d shows that the stages shrink unevenly, so the time moves between them. Vision encoding falls from 33\% to 13\% of the $\pi_{0.5}$ call and prefill rises from 14\% to 34\%, and GR00T N1.7's CPU preprocessing, which no optimization shortens, grows from 12\% to 39\%.

\textbf{On the AMD platform.} Only compilation exists there, so optimization widens the platform gap from 2.0--2.5$\times$ to 5.7--7.8$\times$.

\begin{insightbox}
\textbf{Insight 2.} Optimization makes $\pi_{0.5}$ $7.2\times$ and GR00T N1.7 $2.6\times$ faster on the Jetson, with \textbf{kernel fusion and 4-bit weights giving the largest gain}, and \textbf{moves the remaining time to the stages it shortens least}, such as CPU preprocessing. But fused 4-bit backends exist for only two of the 18 models and only on the Jetson, so \textbf{inference software should bring kernel fusion and low-bit execution to more models and vendors}, and model code should compile as one graph.
\end{insightbox}

\subsection{What Limits the Remaining Time After Optimization?}
\label{sec:q3}

\begin{figure}[t]
  \centering
  \includegraphics[width=\linewidth]{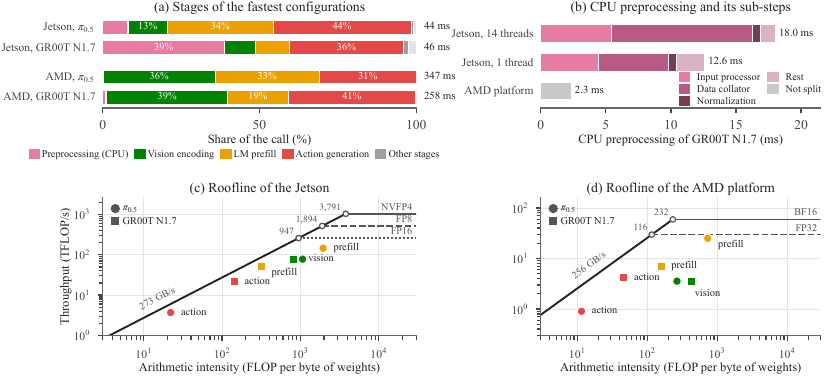}
  \caption{\textbf{What limits the fastest configurations.} (a) Stage shares, with the stages inside a captured graph separated by its kernels. (b) CPU preprocessing of GR00T N1.7 and its sub-steps with PyTorch's 14 CPU threads and with one thread. (c, d) Rooflines, with the peak memory bandwidth as the sloped line, the peak of each number format as a flat line and each GPU stage at its arithmetic intensity $F/W$ and achieved throughput.}
  \label{fig:q3}
  \vspace{-2mm}
\end{figure}

We analyze the fastest configurations of Q2, FlashRT with NVFP4 on the Jetson and compilation on the AMD platform, whose stages Figure~\ref{fig:q3}a shows. Figure~\ref{fig:q3}c,d place the GPU stages on rooflines, and Figure~\ref{fig:q3}b and Appendix Figure~\ref{fig:q3kernels} split them further.

\textbf{Action generation is memory-bound.} A denoising step processes only the 10--41 tokens of the action chunk and robot state, so on the Jetson it performs only 20--150 operations per byte of weights, far below the FP8 and NVFP4 ridge points of 1{,}900 and 3{,}800, and is bounded by reading its weights. FlashRT's steps take only 1.6--1.8$\times$ this bound, as their GEMMs read the weights at three quarters of the peak bandwidth and short normalization, attention and quantization kernels add a third or more to the GEMM time. Action generation on the Jetson is thus limited by the memory bandwidth, a hardware limit that the few action tokens set. On the AMD platform, the steps stay 2.8--3.3$\times$ below the roof, as the compiled GEMMs use less of the bandwidth and Isaac-GR00T recomputes the context's keys and values in every step, which FlashRT does once per call.

\textbf{Prefill length decides its bound.} The 574--712 prefix tokens of $\pi_{0.5}$ put its prefill near or above the ridge. On the AMD platform it is compute-bound, but it reaches only 42\% of the BF16 peak, and a fifth of its tokens are LeRobot's prompt padding. On the Jetson, it lies left of the ridge but stays $3.7\times$ below the roof, as kernels besides the weight GEMMs, such as unfused attention and re-quantization, take half of its time. GR00T N1.7 prefills only 156 tokens and is memory-bound, taking $1.7\times$ its FP8 weight-read time on the Jetson, whereas on the AMD platform its GEMMs read the weights at a fifth of the peak bandwidth. A long prefill is thus limited by its kernels, and a short one by the memory bandwidth on the Jetson and by its GEMMs on the AMD platform.

\textbf{Vision encoding misses the matrix units on the AMD platform.} The vision tower encodes hundreds of patch tokens per image and is compute-bound on the AMD platform, where it is one of the two largest stages of both models, yet runs largely without the matrix units. LeRobot runs $\pi_{0.5}$'s tower in FP32, for which RDNA 3.5 has no matrix instructions, and ROCm's MIOpen executes GR00T N1.7's 3D patch embedding as 512 matrix-vector products without them, so the towers stay $8\times$ and $17\times$ below the roof. On the Jetson, FlashRT's FP8 towers are memory-bound but 2.9--3.7$\times$ below the roof, as normalization, quantization and attention take a third to half of their time. Vision encoding is thus limited by software, by the support for its number format and operators on the AMD platform and by its kernels on the Jetson.

\textbf{CPU preprocessing is limited by host software.} CPU preprocessing is the largest stage of GR00T N1.7 on the Jetson, 39\% of the fastest call, as no optimization touches Isaac-GR00T's input processor and data collator, which prepare images and prompt in Python and PyTorch. Figure~\ref{fig:q3}b shows that a third of it goes to waking PyTorch's 14 worker threads from the deepest idle state their cores enter while the GPU runs, for each small parallel operation, a cost that one thread avoids, and the rest is serial work on one Arm core that the AMD host finishes $5\times$ faster. It is thus limited by host software and its thread configuration, not by the GPU.

\begin{insightbox}
\textbf{Insight 3.} After optimization, \textbf{no single resource limits VLA inference}, as memory bandwidth limits the few-token stages such as the denoising steps on the Jetson, kernels limit vision encoding and prefill, and the host CPU limits preprocessing. \textbf{Edge hardware thus needs bandwidth and a fast CPU more than peak compute}, and \textbf{inference software must fuse the kernels around the GEMMs and streamline preprocessing}, which bandwidth alone cannot fix.
\end{insightbox}

%% file: sections/06_conclusion.tex
\section{Conclusion}
\label{sec:conclusion}


We introduced \bench, which measures {where the inference time of released VLA models goes on edge devices,} stage by stage. Across 18 open models on two edge platforms, latency differs by up to $157\times$ and follows the {number of iterations} of action generation rather than model size; a {backend with kernel fusion and} 4-bit weights gives the largest gain, {2.6--2.9$\times$ over PyTorch with \texttt{torch.compile}}, but exists only on the Jetson and for two models{, and optimization moves time to the stages it shortens least}; and after optimization {memory bandwidth limits only the stages that process few tokens on the Jetson, mainly the denoising steps, while software limits the rest. These insights can guide the design of edge VLA models, inference software and hardware.}

%% file: sections/appendix.tex
\appendix

\section{Additional Benchmark Details}
\label{app:details}

\subsection{Model Configurations}
\label{app:models}

Table~\ref{tab:models-full} complements Table~\ref{tab:models} with the software, the number of camera views, and the chunk size $H$ and the number of actions $H_e$ that the release's own LIBERO evaluation executes per call, for every one of the 19 reference configurations. The $\pi_{0.5}$ and pi0-FAST checkpoints ship with LeRobot's compile option switched on and MolmoAct2 with CUDA graphs, which their reference configurations switch off, and Appendix~\ref{app:q2} reports these defaults.

\begin{table}[H]
  \centering
  \caption{Software, camera views, chunk length $H$ and execution horizon $H_e$ of each configuration.}
  \label{tab:models-full}
  \small
  \setlength{\tabcolsep}{4pt}
  \input{tables/models_full.tex}
\end{table}

{Table~\ref{tab:models-pop} shows how widely the 18 models are used, with the GitHub stars of each code repository and the downloads of the LIBERO checkpoint that we run.}

\begin{table}[H]
  \centering
  \caption{{GitHub stars of the code repository and all-time HuggingFace downloads of the LIBERO checkpoint of each model on 2026-09-26. A dash marks a checkpoint whose downloads HuggingFace does not count.}}
  \label{tab:models-pop}
  \scriptsize
  \setlength{\tabcolsep}{3pt}
  \renewcommand{\UrlBreaks}{\do\/\do\-\do\_\do\.}
  \urlstyle{tt}
  \input{tables/models_popularity.tex}
\end{table}

\subsection{Stage Definitions}
\label{app:stagedefs}

Table~\ref{tab:stagedefs} lists the work in each stage of Equation~\ref{eq:stages}, the unit that executes it, and how it is measured. A stage that a release's code runs inside another one is counted in that one, and Section~\ref{sec:protocol} describes how the stages inside a captured CUDA graph are separated. LeRobot runs vision encoding, prefill and action generation inside one policy call, so we time the policy's own calls. Vision encoding is its image-embedding call, prefill the language-model call that fills the key--value cache, and action generation the rest of the policy call. For FastWAM, vision encoding is the VAE encoding of the image, and prefill comprises the text encoder and the pass of the video transformer over the first frame. In FlashRT's backbone graph of GR00T N1.7, vision encoding ends at the first RMSNorm of the language model, and the VL self-attention after the language model counts as action generation, as in Isaac-GR00T, with each part taking its share of the graph's kernel time.

\begin{table}[H]
  \centering
  \caption{The seven stages of one inference call.}
  \label{tab:stagedefs}
  \small
  \input{tables/stagedefs.tex}
\end{table}

\section{Latency of the Reference Configurations}
\label{app:latency}

Table~\ref{tab:app-latency} lists the numbers behind Figures~\ref{fig:params} and~\ref{fig:q1}a,b. The time per action divides the latency by the chunk size $H$, so a value above 50\,ms means that the configuration generates actions more slowly than LIBERO executes them.

\begin{table}[H]
  \centering
  \caption{Median latency (ms), 5th--95th percentile, inference time per action (ms) and AMD-to-Jetson latency ratio of the 19 reference configurations, sorted by latency on the Jetson.}
  \label{tab:app-latency}
  \small
  \setlength{\tabcolsep}{3.4pt}
  \input{tables/app_latency.tex}
\end{table}

\section{Stage Times}
\label{app:stages}

Tables~\ref{tab:stages-thor} and~\ref{tab:stages-amd} list the stage times of every reference configuration from its instrumented run, sorted by latency on the Jetson. For the five configurations on LeRobot, a second instrumented run with timers at the calls of Appendix~\ref{app:stagedefs} splits the policy call of the first run into vision encoding, prefill and action generation in its own proportions, and its median latency lies within 3.1\% of the first run. An arrow marks a stage that the software fuses into another one.

\begin{table}[H]
  \centering
  \caption{Stage times (ms) on the Jetson AGX Thor.}
  \label{tab:stages-thor}
  \small
  \setlength{\tabcolsep}{4pt}
  \input{tables/stages_thor.tex}
\end{table}

\begin{table}[H]
  \centering
  \caption{Stage times (ms) on the Ryzen AI Max+ PRO 395.}
  \label{tab:stages-amd}
  \small
  \setlength{\tabcolsep}{4pt}
  \input{tables/stages_amd.tex}
\end{table}

\section{Step and View Sweeps}
\label{app:sweeps}

Table~\ref{tab:app-sweeps} lists the linear fits $T = a + bN$ over the number of denoising steps or camera views that give the per-step costs of Section~\ref{sec:q1} and the step costs of the optimized configurations in Sections~\ref{sec:q2} and~\ref{sec:q3}. Repeated sweeps are averaged, and $R^2$ is the lowest among them. A dash marks a configuration that does not exist or was not swept on that platform.

\begin{table}[H]
  \centering
  \caption{Fits $T = a + bN$ of latency over the number of denoising steps or camera views $N$.}
  \label{tab:app-sweeps}
  \small
  \setlength{\tabcolsep}{4pt}
  \input{tables/app_sweeps.tex}
\end{table}

\section{Frame-to-Frame Variation}
\label{app:variation}

Table~\ref{tab:app-variation} lists the spread of the per-frame median over the 24 frames of one episode, relative to the median. For the configurations marked with an asterisk, the values come from the runs of the same model with its default CUDA graphs, because the graph-off configuration was run on the benchmark observation only. This applies to the three MolmoAct2 configurations on the Jetson and to the discrete head on the AMD platform.

\begin{table}[H]
  \centering
  \caption{Spread (\%) of the per-frame median latency over the 24 frames of one episode.}
  \label{tab:app-variation}
  \small
  \setlength{\tabcolsep}{4pt}
  \input{tables/app_variation.tex}
\end{table}

\section{Optimization Details}
\label{app:q2}

Table~\ref{tab:app-q2} lists the latency before and after every optimization we measured, including the options of models outside Section~\ref{sec:q2}, namely the compile option of SmolVLA and pi0-FAST, X-VLA's cast from fp32 to bf16, MolmoAct2's default CUDA graphs and the fast modes of MolmoAct2-Think and DeepThinkVLA. These fast modes reuse or skip reasoning and therefore need their own evaluation of task success. Table~\ref{tab:app-compile} lists every compile mode against the eager reference configuration, and Table~\ref{tab:app-agreement} how much each optimization changes the actions on 11 frames, where the last column is the largest difference between repeated calls of the same configuration on the same frame. The compiled actions of $\pi_{0.5}$ differ by up to 2.0, a flipped gripper command, between two compiled processes on the Jetson and from the eager ones on the AMD platform, far beyond the 0.07 between repeated calls within one process, whereas repeated calls of SmolVLA already differ by up to 1.96. The warm mode of MolmoAct2-Think reuses the depth tokens of the previous frame, so it is timed on the 24 consecutive frames of one episode, one call per frame. FlashRT calibrates its FP8 scales once at load time on the benchmark observation, which is also the first of the 11 frames.

\begin{table}[H]
  \centering
  \caption{Median latency (ms) before and after each optimization of Q2 and the resulting speed-up.}
  \label{tab:app-q2}
  \small
  \setlength{\tabcolsep}{4pt}
  \input{tables/app_q2.tex}
\end{table}

\begin{table}[H]
  \centering
  \caption{Compile modes against the eager reference configuration. GR00T N1.7 compiles only its action head, as in the Isaac-GR00T recipe, and traced frames are the frames that the compiler captures out of those it encounters. The aligned rows run FlashRT's workload of 10 actions and two image slots on both platforms.}
  \label{tab:app-compile}
  \small
  \setlength{\tabcolsep}{4pt}
  \input{tables/app_compile.tex}
\end{table}

\begin{table}[H]
  \centering
  \caption{Agreement of the actions before and after each optimization over 11 frames.}
  \label{tab:app-agreement}
  \small
  \setlength{\tabcolsep}{4pt}
  \input{tables/app_agreement.tex}
\end{table}

\section{Bottleneck Analysis}
\label{app:after}

\begin{figure}[H]
  \centering
  \includegraphics[width=0.62\linewidth]{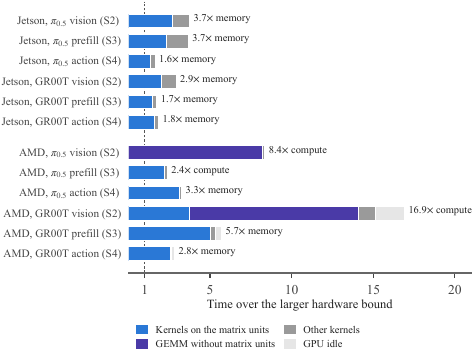}
  \caption{Time of every GPU stage of the fastest configurations over the larger of its two bounds, split by kernel class, with action generation measured per denoising step.}
  \label{fig:q3kernels}
\end{figure}

Figure~\ref{fig:q3kernels} splits the time of every GPU stage of Figure~\ref{fig:q3}c,d into kernel classes, and Table~\ref{tab:app-after} lists the numbers behind these panels. The floating-point operations $F$ and weight bytes $W$ of each stage come from the GEMM shapes of the eager run of the same model and workload and are counted in the number format of each configuration's kernels. FlashRT runs the vision tower and the action expert of $\pi_{0.5}$ with FP8 weights and only the FFN of its language model with NVFP4 weights, runs the vision tower and the language model of GR00T N1.7 with FP8 weights and every GEMM of its action head with NVFP4 weights, and computes the adaptive-norm modulation of the denoising steps and the keys and values of the GR00T N1.7 context once per call, so that a FlashRT step reads only the weights of its GEMMs. The compiled configurations keep the formats of the loaded models, LeRobot pads the $\pi_{0.5}$ prompt to 200 tokens, and Isaac-GR00T recomputes the modulation and the keys and values of the context in every step and computes the output projection of its language model over the whole vocabulary in prefill. The time of a stage inside a captured graph is the kernel time of its part of the trace, or for GR00T N1.7 on FlashRT that part's share of the timed graph, and the time of a GR00T N1.7 step is the slope of its step sweep. The ratio divides the time by the larger of the two bounds, at the peaks of Table~\ref{tab:platforms}.

\begin{table}[H]
  \centering
  \caption{Time of every GPU stage of the fastest configurations against the compute bound $F/P$ and the memory bound $W/B$, with action generation per denoising step.}
  \label{tab:app-after}
  \small
  \setlength{\tabcolsep}{3.5pt}
  \input{tables/app_after.tex}
\end{table}

%% file: tables/models_full.tex
\begin{tabular}{@{}lllccc@{}}
\toprule
Action generation & Model & Software & Views & $H$ & $H_e$ \\
\midrule
Flow matching & X-VLA & authors' code & 2 & 30 & 30 \\
 & GR00T N1.7 & authors' code & 2 & 16 & 8 \\
 & SmolVLA & LeRobot 0.6.1 & 3 & 50 & 50 \\
 & InstructVLA & authors' code & 2 & 8 & 8 \\
 & $\pi_{0.5}$ & LeRobot 0.6.1 & 2 & 50 & 50 \\
 & NORA-1.5 & authors' code & 1 & 5 & 5 \\
 & Evo-1 & authors' code & 2 & 50 & 50 \\
 & FastWAM & LeRobot 0.6.1 & 2 & 32 & 32 \\
 & MolmoAct2 (flow) & LeRobot 0.6.1 & 2 & 10 & 10 \\
 & InternVLA-A1.5 & authors' code & 2 & 8 & 8 \\
\midrule
Parallel & VLA-Adapter & authors' code & 2 & 8 & 8 \\
decoding & SimpleVLA-RL & authors' code & 1 & 8 & 8 \\
 & OpenVLA-OFT & authors' code & 2 & 8 & 8 \\
 & RIPT-VLA & authors' code & 2 & 8 & 8 \\
\midrule
Autoregressive & pi0-FAST & LeRobot 0.6.1 & 2 & 10 & 10 \\
tokens & MolmoAct2 (discrete) & LeRobot 0.6.1 & 2 & 10 & 10 \\
 & VLA-0 & authors' code & 2 & 8 & 1 \\
\midrule
Reasoning, & DeepThinkVLA (CoT) & authors' code & 2 & 10 & 10 \\
then actions & MolmoAct2-Think (cold) & authors' code & 2 & 10 & 10 \\
\bottomrule
\end{tabular}

%% file: tables/models_popularity.tex
\begin{tabular}{@{}l>{\raggedright\arraybackslash}p{4.3cm}>{\raggedright\arraybackslash}p{5.4cm}r@{}}
\toprule
Model & Code repository (stars) & LIBERO checkpoint & Downloads \\
\midrule
X-VLA & \nolinkurl{2toinf/X-VLA} (733) & \nolinkurl{2toINF/X-VLA-Libero} & 12{,}303 \\
GR00T N1.7 & \nolinkurl{NVIDIA/Isaac-GR00T} (8{,}130) & \nolinkurl{nvidia/GR00T-N1.7-LIBERO} & -- \\
SmolVLA & \nolinkurl{huggingface/lerobot} (27{,}783) & \nolinkurl{lerobot/smolvla_libero} & 68{,}467 \\
InstructVLA & \nolinkurl{InternRobotics/InstructVLA} (117) & \nolinkurl{ShuaiYang03/instructvla_pretraining_v2_libero_10_wrist-image_aug} & 220 \\
$\pi_{0.5}$ & \nolinkurl{Physical-Intelligence/openpi} (13{,}997) & \nolinkurl{lerobot/pi05_libero_finetuned_v044} & 174{,}582 \\
NORA-1.5 & \nolinkurl{declare-lab/nora-1.5} (109) & \nolinkurl{declare-lab/nora-1.5-libero} & 109 \\
Evo-1 & \nolinkurl{MINT-SJTU/Evo-1} (364) & \nolinkurl{MINT-SJTU/Evo1_LIBERO} & 823 \\
FastWAM & \nolinkurl{yuantianyuan01/FastWAM} (1{,}526) & \nolinkurl{lerobot/fastwam_libero_uncond_2cam224} & 1{,}813 \\
MolmoAct2 & \nolinkurl{allenai/molmoact2} (772) & \nolinkurl{allenai/MolmoAct2-LIBERO-LeRobot} & 5{,}449 \\
InternVLA-A1.5 & \nolinkurl{InternRobotics/InternVLA-A-series} (558) & \nolinkurl{InternRobotics/InternVLA-A1.5-Libero} & 449 \\
VLA-Adapter & \nolinkurl{OpenHelix-Team/VLA-Adapter} (2{,}314) & \nolinkurl{VLA-Adapter/LIBERO-Long-Pro} & 1{,}171 \\
SimpleVLA-RL & \nolinkurl{PRIME-RL/SimpleVLA-RL} (1{,}865) & \nolinkurl{Haozhan72/openvla-oft-libero10-traj1-rl} & 631 \\
OpenVLA-OFT & \nolinkurl{moojink/openvla-oft} (1{,}399) & \nolinkurl{moojink/openvla-7b-oft-finetuned-libero-10} & 36{,}876 \\
RIPT-VLA & \nolinkurl{Ariostgx/ript-vla} (172) & \nolinkurl{tanshh97/RIPT_VLA} & -- \\
pi0-FAST & \nolinkurl{Physical-Intelligence/openpi} (13{,}997) & \nolinkurl{lerobot/pi0fast-libero} & 68{,}122 \\
VLA-0 & \nolinkurl{NVlabs/vla0} (492) & \nolinkurl{ankgoyal/vla0-libero} & 725 \\
DeepThinkVLA & \nolinkurl{OpenBMB/DeepThinkVLA} (531) & \nolinkurl{yinchenghust/deepthinkvla_libero_cot_rl} & 1{,}052 \\
MolmoAct2-Think & \nolinkurl{allenai/molmoact2} (772) & \nolinkurl{allenai/MolmoAct2-Think-LIBERO} & 1{,}927 \\
\bottomrule
\end{tabular}

%% file: tables/stagedefs.tex
\begin{tabular}{@{}llll@{}}
\toprule
Stage & Work & Unit & Measured by \\
\midrule
S0 Preprocessing & resize, normalize, tokenize & CPU & timer \\
S1 Host-to-device & copy the inputs & copy engine & timer \\
S2 Vision encoding & images to tokens & GPU & timer or kernel trace \\
S3 Prefill & language model, key--value cache & GPU & timer or kernel trace \\
S4 Action generation & denoising steps or decoded tokens & GPU & timer, step sweep or kernel trace \\
S5 Postprocessing & unnormalize, copy back & CPU & timer \\
S6 Residual & latency minus S0--S5 & CPU & derived \\
\bottomrule
\end{tabular}

%% file: tables/app_latency.tex
\begin{tabular}{@{}lrrrrrrr@{}}
\toprule
& \multicolumn{3}{c}{Jetson AGX Thor} & \multicolumn{3}{c}{Ryzen AI Max+ PRO 395} & \\
\cmidrule(lr){2-4}\cmidrule(lr){5-7}
Model & $p_{50}$ & $p_{5}$--$p_{95}$ & Per action & $p_{50}$ & $p_{5}$--$p_{95}$ & Per action & AMD / Jetson \\
\midrule
X-VLA & 76.7 & 75.8--78.8 & 2.6 & 206 & 206--209 & 6.9 & 2.69 \\
VLA-Adapter & 92.7 & 72.8--102 & 11.6 & 114 & 113--115 & 14.2 & 1.23 \\
GR00T N1.7 & 119 & 111--126 & 7.4 & 297 & 296--300 & 18.6 & 2.50 \\
SimpleVLA-RL & 129 & 120--135 & 16.1 & 446 & 444--449 & 55.7 & 3.46 \\
SmolVLA & 191 & 187--195 & 3.8 & 140 & 140--142 & 2.8 & 0.73 \\
InstructVLA & 235 & 228--244 & 29.4 & 357 & 356--367 & 44.7 & 1.52 \\
OpenVLA-OFT & 250 & 243--360 & 31.3 & 495 & 492--498 & 61.8 & 1.98 \\
$\pi_{0.5}$ & 405 & 392--426 & 8.1 & 781 & 778--784 & 15.6 & 1.93 \\
NORA-1.5 & 405 & 399--412 & 81.0 & 335 & 334--338 & 67.0 & 0.83 \\
RIPT-VLA & 419 & 410--435 & 52.3 & 940 & 936--944 & 117 & 2.24 \\
Evo-1 & 453 & 442--467 & 9.1 & 711 & 710--713 & 14.2 & 1.57 \\
FastWAM & 523 & 515--529 & 16.4 & 1101 & 1089--1119 & 34.4 & 2.10 \\
MolmoAct2 (flow) & 553 & 544--565 & 55.3 & 459 & 457--462 & 45.9 & 0.83 \\
pi0-FAST & 989 & 985--993 & 98.9 & 2497 & 2493--2502 & 250 & 2.52 \\
InternVLA-A1.5 & 1232 & 1177--1296 & 154 & 830 & 809--890 & 104 & 0.67 \\
MolmoAct2 (discrete) & 1275 & 1271--1281 & 128 & 1830 & 1828--1831 & 183 & 1.43 \\
DeepThinkVLA (CoT) & 2396 & 2384--2411 & 240 & 5828 & 5821--5836 & 583 & 2.43 \\
MolmoAct2-Think (cold) & 4457 & 4441--4464 & 446 & 6188 & 6185--6240 & 619 & 1.39 \\
VLA-0 & 6293 & 6244--6310 & 787 & 17889 & 17877--17896 & 2236 & 2.84 \\
\bottomrule
\end{tabular}

%% file: tables/stages_thor.tex
\begin{tabular}{@{}lrrrrrrrr@{}}
\toprule
Model & S0 & S1 & S2 & S3 & S4 & S5 & S6 & Total \\
\midrule
X-VLA & 6.0 & 0.9 & $\rightarrow$ & 19.1 & 51.7 & 0.1 & 0.4 & 78 \\
VLA-Adapter & 22.8 & 0.8 & 27.0 & 20.8 & 21.7 & 2.4 & 0.8 & 96 \\
GR00T N1.7 & 14.2 & 1.0 & 19.6 & 19.4 & 67.7 & 0.2 & 1.2 & 123 \\
SimpleVLA-RL & 20.1 & 0.9 & 23.5 & 85.7 & $\rightarrow$ & 2.2 & -3.0 & 129 \\
SmolVLA & 4.3 & $\rightarrow$ & 20.9 & 15.0 & 146.6 & 0.1 & 0.2 & 187 \\
InstructVLA & 9.6 & $\rightarrow$ & 24.8 & 36.2 & 149.1 & 17.5 & 0.7 & 238 \\
OpenVLA-OFT & 49.3 & 1.1 & 44.0 & 154.6 & 2.1 & 3.0 & 0.2 & 254 \\
$\pi_{0.5}$ & 3.0 & $\rightarrow$ & 163.1 & 69.4 & 168.1 & 0.1 & -0.2 & 403 \\
NORA-1.5 & 3.8 & $\rightarrow$ & 34.7 & 47.4 & 319.0 & 0.1 & 4.5 & 410 \\
RIPT-VLA & 26.4 & 0.9 & 71.6 & 223.6 & 99.5 & 2.5 & -1.1 & 423 \\
Evo-1 & 7.9 & 0.6 & $\rightarrow$ & 261.1 & 219.5 & 0.2 & 1.4 & 491 \\
FastWAM & 3.5 & $\rightarrow$ & 18.2 & 137.0 & 361.7 & 0.2 & 0.2 & 521 \\
MolmoAct2 (flow) & 23.2 & $\rightarrow$ & 38.2 & 100.4 & 392.8 & 0.3 & -1.8 & 553 \\
pi0-FAST & 5.7 & $\rightarrow$ & 168.4 & 66.2 & 774.1 & 1.3 & 0.1 & 1016 \\
InternVLA-A1.5 & 16.4 & 0.5 & 22.1 & 127.8 & 1066.1 & 0.2 & 2.9 & 1236 \\
MolmoAct2 (discrete) & 26.2 & $\rightarrow$ & 33.8 & 87.6 & 1131.4 & 0.4 & -0.5 & 1279 \\
DeepThinkVLA (CoT) & 6.0 & 0.6 & 24.3 & 53.2 & 2186.3 & 172.4 & 0.0 & 2443 \\
MolmoAct2-Think (cold) & 20.5 & $\rightarrow$ & 33.1 & 95.3 & 4436.1 & 0.0 & 49.7 & 4635 \\
VLA-0 & 11.3 & $\rightarrow$ & 47.9 & $\rightarrow$ & 6235.8 & 0.2 & 1.9 & 6297 \\
\bottomrule
\end{tabular}

%% file: tables/stages_amd.tex
\begin{tabular}{@{}lrrrrrrrr@{}}
\toprule
Model & S0 & S1 & S2 & S3 & S4 & S5 & S6 & Total \\
\midrule
X-VLA & 2.8 & 0.4 & $\rightarrow$ & 51.5 & 154.3 & 0.1 & 0.1 & 209 \\
VLA-Adapter & 5.9 & 0.3 & 60.2 & 29.7 & 15.8 & 1.8 & 0.1 & 114 \\
GR00T N1.7 & 2.4 & 0.9 & 98.6 & 49.6 & 145.3 & 0.1 & 0.7 & 298 \\
SimpleVLA-RL & 3.3 & 0.3 & 32.7 & 408.4 & $\rightarrow$ & 1.0 & -0.0 & 446 \\
SmolVLA & 0.7 & $\rightarrow$ & 40.8 & 10.9 & 87.5 & 0.1 & 0.0 & 140 \\
InstructVLA & 3.8 & $\rightarrow$ & 49.5 & 71.8 & 224.6 & 7.5 & 0.5 & 358 \\
OpenVLA-OFT & 6.0 & 0.3 & 62.2 & 423.4 & 3.5 & 1.2 & 0.2 & 497 \\
$\pi_{0.5}$ & 0.7 & $\rightarrow$ & 286.0 & 278.1 & 215.9 & 0.1 & 0.2 & 781 \\
NORA-1.5 & 1.0 & $\rightarrow$ & 56.3 & 99.3 & 175.2 & 0.1 & 2.9 & 335 \\
RIPT-VLA & 6.3 & 0.3 & 126.9 & 593.8 & 210.6 & 1.7 & 0.4 & 940 \\
Evo-1 & 4.3 & 1.0 & $\rightarrow$ & 362.6 & 342.7 & 0.4 & 0.1 & 711 \\
FastWAM & 1.4 & $\rightarrow$ & 50.2 & 402.4 & 648.3 & 0.2 & 0.3 & 1103 \\
MolmoAct2 (flow) & 7.4 & $\rightarrow$ & 58.4 & 156.8 & 236.5 & 0.4 & 0.2 & 460 \\
pi0-FAST & 1.9 & $\rightarrow$ & 286.9 & 272.1 & 1936.5 & 0.6 & 0.1 & 2498 \\
InternVLA-A1.5 & 4.5 & 0.8 & 97.4 & 158.2 & 569.0 & 0.1 & 7.0 & 837 \\
MolmoAct2 (discrete) & 6.7 & $\rightarrow$ & 58.3 & 157.3 & 1608.3 & 0.2 & 0.0 & 1831 \\
DeepThinkVLA (CoT) & 2.6 & 0.7 & 42.0 & 186.5 & 5552.0 & 52.3 & 1.0 & 5837 \\
MolmoAct2-Think (cold) & 6.4 & $\rightarrow$ & 58.9 & 165.6 & 5945.7 & 0.0 & 38.5 & 6215 \\
VLA-0 & 3.4 & $\rightarrow$ & 92.6 & $\rightarrow$ & 17795.7 & 0.1 & 0.4 & 17892 \\
\bottomrule
\end{tabular}

%% file: tables/app_sweeps.tex
\begin{tabular}{@{}llrrrrrr@{}}
\toprule
& & \multicolumn{3}{c}{Jetson AGX Thor} & \multicolumn{3}{c}{Ryzen AI Max+ PRO 395} \\
\cmidrule(lr){3-5}\cmidrule(lr){6-8}
Configuration & Swept & $b$ (ms) & $a$ (ms) & $R^2$ & $b$ (ms) & $a$ (ms) & $R^2$ \\
\midrule
\multicolumn{8}{@{}l}{\textit{Reference configurations}} \\
X-VLA & steps & 4.9 & 29 & 0.999 & 15.2 & 56 & 1.000 \\
GR00T N1.7 & steps & 15.7 & 53 & 0.996 & 34.2 & 161 & 1.000 \\
SmolVLA & steps & 14.5 & 41 & 1.000 & 8.6 & 54 & 1.000 \\
InstructVLA & steps & 13.3 & 103 & 1.000 & 19.5 & 163 & 1.000 \\
$\pi_{0.5}$ & steps & 14.7 & 259 & 0.999 & 21.3 & 568 & 1.000 \\
NORA-1.5 & steps & 31.4 & 86 & 0.999 & 17.6 & 159 & 1.000 \\
Evo-1 & steps & 3.9 & 280 & 0.988 & 6.8 & 371 & 1.000 \\
FastWAM & steps & 32.7 & 190 & 0.999 & 64.4 & 463 & 1.000 \\
MolmoAct2 (flow) & steps & 37.8 & 162 & 0.999 & 21.7 & 242 & 1.000 \\
InternVLA-A1.5 & steps & 106.8 & 176 & 1.000 & 55.9 & 268 & 1.000 \\
MolmoAct2-Think (cold) & steps & 34.0 & 4128 & 0.989 & 22.2 & 5993 & 0.999 \\
\midrule
\multicolumn{8}{@{}l}{\textit{Paths of Q2}} \\
X-VLA in FP32 & steps & 51.9 & 127 & 1.000 & 74.9 & 175 & 1.000 \\
SmolVLA, compile (reduce-overhead) & steps & 2.7 & 40 & 0.986 & 5.9 & 50 & 1.000 \\
SmolVLA, compile (max-autotune) & steps & 2.4 & 41 & 0.986 & -- & -- & -- \\
$\pi_{0.5}$, compile (max-autotune, shipped) & steps & 6.8 & 90 & 1.000 & 8.4 & 358 & 0.999 \\
$\pi_{0.5}$, compile (reduce-overhead) & steps & 6.5 & 91 & 1.000 & 17.3 & 507 & 1.000 \\
MolmoAct2 (flow), CUDA graphs on & steps & 10.9 & 178 & 0.984 & -- & -- & -- \\
GR00T N1.7, TensorRT & steps & 11.8 & 49 & 0.999 & -- & -- & -- \\
GR00T N1.7, FlashRT FP16 & steps & 10.0 & 38 & 0.868 & -- & -- & -- \\
$\pi_{0.5}$, FlashRT FP16 & views & 12.6 & 48 & 1.000 & -- & -- & -- \\
\bottomrule
\end{tabular}

%% file: tables/app_variation.tex
\begin{tabular}{@{}lrr@{}}
\toprule
Model & Jetson AGX Thor & Ryzen AI Max+ PRO 395 \\
\midrule
X-VLA & 3 & 1 \\
VLA-Adapter & 25 & 1 \\
GR00T N1.7 & 15 & 0 \\
SimpleVLA-RL & 10 & 0 \\
SmolVLA & 5 & 0 \\
InstructVLA & 4 & 0 \\
OpenVLA-OFT & 10 & 0 \\
$\pi_{0.5}$ & 2 & 0 \\
NORA-1.5 & 4 & 0 \\
RIPT-VLA & 3 & 0 \\
Evo-1 & 3 & 0 \\
FastWAM & 7 & 1 \\
MolmoAct2 (flow)$^\ast$ & 5 & 0 \\
pi0-FAST & 38 & 42 \\
InternVLA-A1.5 & 7 & 3 \\
MolmoAct2 (discrete)$^\ast$ & 122 & 121 \\
DeepThinkVLA (CoT) & 20 & 19 \\
MolmoAct2-Think (cold)$^\ast$ & 1 & 0 \\
VLA-0 & 12 & 11 \\
\bottomrule
\end{tabular}

%% file: tables/app_q2.tex
\begin{tabular}{@{}lrrrrrr@{}}
\toprule
& \multicolumn{3}{c}{Jetson AGX Thor} & \multicolumn{3}{c}{Ryzen AI Max+ PRO 395} \\
\cmidrule(lr){2-4}\cmidrule(lr){5-7}
Optimization & Before & After & Speed-up & Before & After & Speed-up \\
\midrule
SmolVLA, compile (reduce-overhead) & 191 & 66.4 & 2.88 & 140 & 109 & 1.28 \\
SmolVLA, compile (max-autotune) & 191 & 71.9 & 2.66 & 140 & 77.7 & 1.80 \\
$\pi_{0.5}$, compile (reduce-overhead) & 405 & 157 & 2.58 & 781 & 682 & 1.14 \\
$\pi_{0.5}$, compile (max-autotune, shipped) & 405 & 159 & 2.55 & 781 & 442 & 1.77 \\
$\pi_{0.5}$, compile (max-autotune), FlashRT's workload & 318 & 128 & 2.48 & 631 & 347 & 1.82 \\
GR00T N1.7, compile recipe (Isaac-GR00T) & 119 & 117 & 1.02 & 297 & 258 & 1.15 \\
pi0-FAST, compile (reduce-overhead) & 989 & 874 & 1.13 & 2497 & 2498 & 1.00 \\
MolmoAct2 (flow), CUDA graphs & 553 & 288 & 1.92 & -- & -- & -- \\
MolmoAct2 (discrete), CUDA graphs & 1275 & 1129 & 1.13 & 1830 & 1612 & 1.14 \\
MolmoAct2-Think, CUDA graphs & 4457 & 4191 & 1.06 & -- & -- & -- \\
\midrule
$\pi_{0.5}$, eager LeRobot $\rightarrow$ FlashRT & 405 & 72.8 & 5.56 & -- & -- & -- \\
$\pi_{0.5}$, compiled LeRobot $\rightarrow$ FlashRT & 157 & 72.8 & 2.15 & -- & -- & -- \\
$\pi_{0.5}$, compiled LeRobot $\rightarrow$ FlashRT, same workload & 128 & 72.8 & 1.76 & -- & -- & -- \\
GR00T N1.7, PyTorch $\rightarrow$ TensorRT & 119 & 96.2 & 1.24 & -- & -- & -- \\
GR00T N1.7, PyTorch $\rightarrow$ FlashRT & 119 & 80.7 & 1.47 & -- & -- & -- \\
\midrule
X-VLA, FP32 $\rightarrow$ BF16 & 646 & 76.7 & 8.42 & 922 & 206 & 4.47 \\
$\pi_{0.5}$, FP16 $\rightarrow$ FP8 & 72.8 & 47.9 & 1.52 & -- & -- & -- \\
$\pi_{0.5}$, FP16 $\rightarrow$ NVFP4 & 72.8 & 44.4 & 1.64 & -- & -- & -- \\
GR00T N1.7, FP16 $\rightarrow$ FP8 & 80.7 & 60.9 & 1.33 & -- & -- & -- \\
GR00T N1.7, FP16 $\rightarrow$ NVFP4 & 80.7 & 45.6 & 1.77 & -- & -- & -- \\
\midrule
MolmoAct2-Think, reuse depth tokens & 4191 & 1541 & 2.72 & 6188 & 2355 & 2.63 \\
DeepThinkVLA, mask the reasoning & 2396 & 142 & 16.84 & 5828 & 254 & 22.97 \\
\bottomrule
\end{tabular}

%% file: tables/app_compile.tex
\begin{tabular}{@{}lllrrr@{}}
\toprule
Model & Platform & Mode & $p_{50}$ (ms) & Speed-up & Traced frames \\
\midrule
SmolVLA & Jetson & reduce-overhead & 66.4 & 2.88 & 14/17 \\
SmolVLA & AMD & reduce-overhead & 109 & 1.28 & -- \\
SmolVLA & Jetson & max-autotune & 71.9 & 2.66 & -- \\
SmolVLA & AMD & max-autotune & 77.7 & 1.80 & -- \\
\midrule
$\pi_{0.5}$ & Jetson & max-autotune (shipped) & 159 & 2.55 & -- \\
$\pi_{0.5}$ & AMD & max-autotune (shipped) & 442 & 1.77 & -- \\
$\pi_{0.5}$ & Jetson & reduce-overhead & 157 & 2.58 & 6/6 \\
$\pi_{0.5}$ & AMD & reduce-overhead & 682 & 1.14 & -- \\
$\pi_{0.5}$ & Jetson & max-autotune, aligned & 128 & 2.48 & 6/6 \\
$\pi_{0.5}$ & AMD & max-autotune, aligned & 347 & 1.82 & 1/1 \\
\midrule
pi0-FAST & Jetson & reduce-overhead & 874 & 1.13 & 0/0 \\
pi0-FAST & AMD & reduce-overhead & 2498 & 1.00 & 0/0 \\
\midrule
GR00T N1.7 & Jetson & max-autotune, action head & 117 & 1.02 & 1/1 \\
GR00T N1.7 & AMD & max-autotune, action head & 258 & 1.15 & 1/1 \\
\bottomrule
\end{tabular}

%% file: tables/app_agreement.tex
\begin{tabular}{@{}llrrrr@{}}
\toprule
& & \multicolumn{2}{c}{Cosine similarity} & \multicolumn{2}{c}{Max $|\Delta|$} \\
\cmidrule(lr){3-4}\cmidrule(lr){5-6}
Optimization & Platform & min & median & optimized & repeated calls \\
\midrule
$\pi_{0.5}$, FlashRT FP16 $\rightarrow$ FP8 & Jetson & 0.99839 & 0.99938 & 0.124 & 0.059 \\
$\pi_{0.5}$, FlashRT FP16 $\rightarrow$ NVFP4 & Jetson & 0.93429 & 0.99812 & 0.677 & 0.096 \\
GR00T N1.7, FlashRT FP16 $\rightarrow$ FP8 & Jetson & 0.99986 & 0.99993 & 0.033 & 0.003 \\
GR00T N1.7, FlashRT FP16 $\rightarrow$ NVFP4 & Jetson & 0.99974 & 0.99986 & 0.047 & 0.003 \\
X-VLA, FP32 $\rightarrow$ BF16 & Jetson & 0.99999 & 0.99999 & 0.004 & 0.004 \\
SmolVLA, eager $\rightarrow$ compile (max-autotune) & Jetson & 0.96950 & 0.99931 & 1.918 & 0.251 \\
pi0-FAST, eager $\rightarrow$ compile (max-autotune) & Jetson & 1.00000 & 1.00000 & 0.000 & 0.000 \\
$\pi_{0.5}$, two compiled processes (max-autotune) & Jetson & 0.96649 & 0.99958 & 2.000 & 0.054 \\
SmolVLA, eager $\rightarrow$ compile (reduce-overhead) & Jetson & 0.97089 & 0.99915 & 1.865 & 1.870 \\
$\pi_{0.5}$, max-autotune $\rightarrow$ reduce-overhead & Jetson & 0.96659 & 0.99952 & 1.995 & 0.070 \\
pi0-FAST, eager $\rightarrow$ compile (reduce-overhead) & Jetson & 1.00000 & 1.00000 & 0.000 & 0.000 \\
MolmoAct2 (flow), CUDA graphs & Jetson & 0.99979 & 0.99996 & 0.038 & 0.016 \\
MolmoAct2 (discrete), CUDA graphs & Jetson & 1.00000 & 1.00000 & 0.000 & 0.000 \\
MolmoAct2-Think, CUDA graphs & Jetson & 1.00000 & 1.00000 & 0.000 & 0.000 \\
$\pi_{0.5}$, eager $\rightarrow$ compile (max-autotune, shipped) & Jetson & 0.99928 & 0.99960 & 0.083 & 0.054 \\
GR00T N1.7, eager $\rightarrow$ compile recipe & Jetson & 0.99926 & 0.99985 & 0.071 & 0.038 \\
X-VLA, FP32 $\rightarrow$ BF16 & AMD & 0.99999 & 0.99999 & 0.005 & 0.004 \\
SmolVLA, eager $\rightarrow$ compile (max-autotune) & AMD & 0.93653 & 0.99929 & 2.004 & 1.956 \\
SmolVLA, eager $\rightarrow$ compile (reduce-overhead) & AMD & 0.91342 & 0.99935 & 2.012 & 1.961 \\
$\pi_{0.5}$, max-autotune $\rightarrow$ reduce-overhead & AMD & 0.96675 & 0.99957 & 1.992 & 0.050 \\
pi0-FAST, eager $\rightarrow$ compile (reduce-overhead) & AMD & 1.00000 & 1.00000 & 0.000 & 0.000 \\
MolmoAct2 (discrete), CUDA graphs & AMD & 1.00000 & 1.00000 & 0.000 & 0.000 \\
$\pi_{0.5}$, eager $\rightarrow$ compile (max-autotune, shipped) & AMD & 0.96725 & 0.99965 & 2.000 & 0.050 \\
GR00T N1.7, eager $\rightarrow$ compile recipe & AMD & 0.99965 & 0.99989 & 0.047 & 0.033 \\
$\pi_{0.5}$, eager $\rightarrow$ compile (max-autotune, cold cache) & AMD & 0.96639 & 0.99960 & 1.993 & 0.035 \\
pi0-FAST, eager $\rightarrow$ compile (max-autotune) & AMD & 1.00000 & 1.00000 & 0.000 & 0.000 \\
\bottomrule
\end{tabular}

%% file: tables/app_after.tex
\begin{tabular}{@{}lllrrrrrr@{}}
\toprule
Configuration & Stage & Format & Time & $F$ & $W$ & $F/P$ & $W/B$ & Ratio \\
 & & & (ms) & (GFLOP) & (MB) & (ms) & (ms) & \\
\midrule
Jetson, $\pi_{0.5}$ & Vision (S2) & FP8 & 5.66 & 439.7 & 413 & 0.88 & 1.51 & 3.7 \\
 & Prefill (S3) & NVFP4, FP8 & 15.20 & 2205.4 & 1132 & 2.45 & 4.15 & 3.7 \\
 & Action (S4) & FP8 & 1.92 & 7.1 & 322 & 0.02 & 1.18 & 1.6 \\
\addlinespace[2pt]
Jetson, GR00T N1.7 & Vision (S2) & FP8 & 4.61 & 349.5 & 431 & 0.72 & 1.58 & 2.9 \\
 & Prefill (S3) & FP8 & 5.12 & 254.4 & 805 & 0.50 & 2.95 & 1.7 \\
 & Action (S4) & NVFP4 & 3.30 & 69.8 & 489 & 0.07 & 1.79 & 1.8 \\
\addlinespace[2pt]
AMD, $\pi_{0.5}$ & Vision (S2) & FP32 & 123.64 & 439.7 & 1654 & 14.81 & 6.46 & 8.4 \\
 & Prefill (S3) & BF16 & 115.92 & 2896.8 & 3964 & 48.78 & 15.48 & 2.4 \\
 & Action (S4) & BF16 & 8.09 & 7.3 & 636 & 0.12 & 2.49 & 3.3 \\
\addlinespace[2pt]
AMD, GR00T N1.7 & Vision (S2) & BF16 & 99.51 & 349.5 & 808 & 5.88 & 3.16 & 16.9 \\
 & Prefill (S3) & BF16 & 49.79 & 351.5 & 2233 & 5.92 & 8.72 & 5.7 \\
 & Action (S4) & BF16 & 24.21 & 101.4 & 2199 & 1.71 & 8.59 & 2.8 \\
\bottomrule
\end{tabular}